\documentclass[screen,authorversion]{acmart} % nonacm

\usepackage{color}
\usepackage{csquotes}
\usepackage{enumitem}
\usepackage[nolist, nohyperlinks]{acronym}

\acrodefplural{MRS}[MRSs]{Music Recommender Systems}
\acrodefplural{RS}[RSs]{Recommender Systems}
\acrodefplural{GPU}[GPUs]{Graphics Processing Units}
\begin{acronym}
    \acro{MRS}{Music Recommender System}
    \acro{RS}{Recommender System}
    \acro{XAI}{eXplainable Artificial Intelligence}
    \acro{GPU}{Graphics Processing Unit}
    \acro{KG}{Knowledge Graph}
    \acro{ML}{Machine Learning}
\end{acronym}

\newcommand{\ie}{i.\,e., }
\newcommand{\eg}{e.\,g., }
\newcommand{\wrt}{w.\,r.\,t. }

\newcommand{\vs}{vs. }
\newcommand{\examp}[1]{\textit{"#1"}}
\newcommand{\quotee}[1]{\textit{"#1"}}

\newcommand{\optcite}[1]{\cite{#1}} %cite optional ref
\newcommand{\multipleoptcite}[2]{\cite{#1,#2}} %cite both non optional and optional references.
\AtBeginDocument{%
  \providecommand\BibTeX{{%
    \normalfont B\kern-0.5em{\scshape i\kern-0.25em b}\kern-0.8em\TeX}}}

\acmBooktitle{AI Magazine 2022}
\setcopyright{none}
\begin{document}

\title{Explainability in Music Recommender Systems}
%\titlenote{To appear in AI Magazine, Special Topic on Recommender Systems 2022} % subtitle


\author{Darius Afchar}
\orcid{0000-0002-4315-1461}
\authornote{Both authors contributed equally to this research.}
\email{research@deezer.com}
\affiliation{
\institution{Deezer Research}
\country{Paris, France}
}

\author{Alessandro B. Melchiorre}
\email{alessandro.melchiorre@jku.at}
\orcid{0000-0003-1643-1166}
\authornotemark[1]
\author{Markus Schedl}
\orcid{0000-0003-1706-3406}
\email{markus.schedl@jku.at}
\affiliation{
\institution{Johannes Kepler University and Linz Institute of Technology}
\country{Linz, Austria}
}

\author{Romain Hennequin}
\email{research@deezer.com}
\author{Elena V. Epure}
\email{research@deezer.com}
\author{Manuel Moussallam}
\email{research@deezer.com}
\affiliation{
\institution{Deezer Research}
\country{Paris, France}
}

\renewcommand{\shortauthors}{Afchar et al.}

\begin{abstract} % [Abstract of 250 words.]
The most common way to listen to recorded music nowadays is via streaming platforms which provide access to tens of millions of tracks. To assist users in effectively browsing these large catalogs, the integration of Music Recommender Systems (MRSs) has become essential. Current real-world MRSs are often quite complex and optimized for recommendation accuracy. They combine several building blocks based on collaborative filtering and content-based recommendation. This complexity can hinder the ability to explain recommendations to end users, which is particularly important for recommendations perceived as unexpected or inappropriate. While pure recommendation performance often correlates with user satisfaction, explainability has a positive impact on other factors such as trust and forgiveness, which are ultimately essential to maintain user loyalty.

In this article, we discuss how explainability can be addressed in the context of MRSs. 
We provide perspectives on how explainability could improve music recommendation algorithms and enhance user experience. 
First, we review common dimensions and goals of recommenders' explainability and in general of eXplainable Artificial Intelligence (XAI),
and elaborate on the extent to which these apply -- or need to be adapted -- to the specific characteristics of music consumption and recommendation.
Then, we show how explainability components can be integrated within a MRS and in what form explanations can be provided. 
Since the evaluation of explanation quality is decoupled from pure accuracy-based evaluation criteria, we also discuss requirements and strategies for evaluating explanations of music recommendations.
Finally, we describe the current challenges for introducing explainability within a large-scale industrial music recommender system and provide research perspectives.
\end{abstract}

\maketitle

\section{Music Recommender Systems}\label{sec:MRS}
\ac{RS} technology permeates our daily lives; whether we are looking for a book to buy, movie to watch, or accommodation for our next vacation, \acp{RS} are omnipresent.
Like \acp{RS} in other domains~\cite{DBLP:reference/sp/2015rsh}, \acp{MRS}~\cite{DBLP:reference/sp/SchedlKMBK15} have information filtering algorithms at their core, which select from a commonly huge catalog of music items (\eg~artists, albums, or songs) those 
%items the algorithm identifies
identified as most relevant %appropriate, , or pleasant 
for a target user.
%As such, 
Thus \acp{MRS} 
%provide their users guidance
guide users in the otherwise sheer overwhelming amount of music available at their fingertips nowadays.\footnote{The catalogs of music streaming platforms such as Deezer, Spotify, or Pandora include several tens of million music pieces.}

The raising awareness of, and ongoing discussion about, %fairness and 
transparency of machine learning algorithms, including those used in \acp{RS}, has resulted in a substantial demand from users to receive explanations for why certain items have been recommended to them~\cite{DBLP:journals/ftir/ZhangC20}.
% The research area dealing, among others, with explainability aspects in machine learning is frequently referred to as \ac{XAI}.
Also from a \ac{RS} provider's perspective, these aspects are important for building and maintaining trust of the users in the system.
Therefore, equipping \acp{MRS} with capabilities to provide explanations to their users is of mutual interest. %; and its corresponding research a timely topic.

\subsection{Characteristics of music consumption and music recommender systems}\label{sec:specificities}
While %the task of 
music recommendation shares some properties with other media recommendation tasks, such as %user-generated
videos or movies, there exist also pronounced differences. Among the ones identified in literature (\eg~\cite{DBLP:journals/ijmir/SchedlZCDE18}), the following characteristics are relevant for explainability in \acp{MRS}, as we will elaborate in the subsequent sections:

\begin{itemize} 
\item The \textit{duration} of item consumption is commonly much shorter than in other domains, \ie~songs have typical lengths of several minutes, whereas %videos or 
watching a movie, reading a book, or spending a holiday take much longer. % time to consume. %are much longer. %Together with the huge amount of music nowadays available to consumers on streaming platforms, music has become more ``disposable'' than in the pre-streaming era; and listeners need to be convinced in a much shorter time of a recommended song. These facts even changed the way music is produced: shorter duration and earlier chorus.\footnote{https://www.octiive.com/blog/streaming-changing-music-production}

\item Music data comes in \textit{manifold representations}, including audio, % score sheet, 
%symbolic 
MIDI, and textual metadata (\eg~editorial metadata %such as artist, track name, album release year, and genre, 
but also user-generated tags). Furthermore, music-related data that can be leveraged in \acp{MRS} is highly multimodal and includes images (\eg~album covers) and videos (\eg~music video clips), next to audio and textual metadata. Finally, %in a recommender context, 
%collaborative knowledge 
user feedback is collected from various activities (\eg \textit{likes}, \textit{favorites}, \textit{song skips}). %, \textit{searches},  ...) 
%and by matching of user profiles and communities (\eg \textit{age}, \textit{personality}).

\item The listening \textit{context} strongly affects music preferences \optcite{ibrahim2020should}. 
For instance, the listener's mood, location, (\eg~consumption at home \vs~while commuting), 
social situation (\eg~alone \vs~with friends) and other aspects 
have been shown to influence musical needs and demands~\cite{Rentfrow_etal:JP:2011,DBLP:conf/um/FerwerdaST15}. %Context-awareness is therefore of particular importance in \acp{MRS}.

%\item Music listening behavior has different \textit{attention levels}.
%Users can listen to music passively in which case they seek a lean-back experience and no interaction with the \ac{MRS}. Alternatively, they could actively engage with the \ac{MRS}, \ie~seek a lean-in experience, aiming to actively discover new music or to create a playlist, which can be supported by song recommendations.

\item Music is often consumed sequentially, \ie~tracks in a listening session or playlist. Therefore, for music, we often focus on {sequential recommendation} tasks, such as automatic playlist creation or continuation~\cite{DBLP:journals/csur/BonninJ14,DBLP:journals/tist/ZamaniSLC19},
that leverage both long-term and short-term user preferences.

%\item The same songs are often \textit{consumed repeatedly}. %item (music piece)  \textit{Repeated consumption} of is common in \acp{MRS}.
%and is necessary for absorption.

%\item Recommendations are supposed to be \textit{immediately consumed} by the user.
%of music". As music recommendations are intended to be consumed immediately by listeners and therefore the system should be reactive and responsive to the listener's feedback
 
%\item When modeling user preferences, \textit{temporality} is an important factor. Therefore, music RSs often leverage both long-term interest and short-term preferences of their users.
%Considering temporal aspects in the recommendation are also important when users strive to re-discover music that they have not listened to for a while.

\end{itemize}

\subsection{Common music recommendation tasks and methods}
Various use cases of \acp{MRS} exist, centered around different tasks. 
Among these, the most important ones are \textit{front page recommendation} (recommending content for thematic collections of music -- also known as shelves or channels -- presented to the user on the front page of the platform's user interface)~\cite{bendada2020carousel}, \textit{music exploration/discovery} (\eg~based on item similarity in terms of melody, rhythm, or lyrics)~\cite{DBLP:journals/tismir/KneesSG20,DBLP:journals/spm/GotoD19}, \textit{automatic playlist generation} (commonly based on the user profile, but possibly only based on a seed description such as ``music to relax''), and \textit{automatic playlist continuation} (based on a sequence of seed tracks)~\cite{DBLP:conf/recsys/JannachLK15,DBLP:journals/tist/ZamaniSLC19}.

To create a music recommendation engine, a variety of methods are adopted, depending on the use case. These include latent factor models (\eg~singular value decomposition~\optcite{DBLP:journals/eswa/KimY05} or factorization machines~\optcite{DBLP:conf/ecir/LoniSLH14}), graph mining techniques (\eg~random walks~\optcite{DBLP:conf/ijcai/GoriP07} or graph embeddings~\optcite{DBLP:journals/eswa/PalumboMRTB20}), and deep learning-based techniques (\eg~convolutional neural networks~\optcite{DBLP:conf/nips/OordDS13}, recurrent neural networks~\optcite{DBLP:journals/corr/HidasiKBT15}, or autoencoders~\optcite{DBLP:conf/www/LiangKHJ18}). Furthermore, techniques from audio signal processing and natural language processing are often used to create vector representations of music items or to annotate music items with relevant tags \optcite{oramas2018natural, epure2020multilingual,ibrahim2020audio}. % cite oramas2018natural

%Production-ready music \acp{RS} are usually made of several building blocks that take advantage of various methods and sources of music data. This opens many possibilities to create explanations for items suggested by \acp{MRS}.

\vspace{1em}
In this article, we discuss how explainability can be approached in \acp{MRS}, and we provide perspectives and outline challenges in this context. More precisely, we first review definitions and goals of explainability commonly adopted in \ac{RS} research, and investigate to which extent they are applicable or need adaptation in the music domain (Section~\ref{sec:goals}).
Subsequently, Section~\ref{sec:explainabilityMRS} reviews existing explanation types and describes the means through which explanations can be provided to the users, and the methods to integrate explainability capabilities into \acp{MRS}.
How to evaluate the offered explanations in a music recommendation context is discussed in Section~\ref{sec:evaluation}.
Finally, taking an industry perspective, Section~\ref{sec:challenges_industry} describes the challenges \ac{MRS} providers face when integrating explainability functionality into their real-world systems.

\section{Goals and Dimensions of Explainability for Music Recommender Systems} 
\label{sec:goals}

Recent years have seen an upsurge of interest in explainable recommendations, even though the concept already emerged in the 2000s \cite{schafer1999recommender}. % CITE: schafer1999recommender
This evolution of explainable \acp{RS} has been accompanied by an increasing popularity of \acf{XAI}, with which it shares roots, approaches, and terminology. \ac{XAI} represents the convergence of many research disciplines, including computer science, human-computer-interaction, philosophy, and psychology. Coherent and stable XAI definitions and terms have started to appear only recently~\cite{arrieta2020explainable,guidotti2018survey,lipton2018mythos}. Meanwhile, \acp{RS} research has developed explanation-related concepts that are unknown to general \ac{XAI}, some of which, however, rest upon elusive descriptions. This specificity is probably due to the nature of \acp{RS} themselves, which differ \wrt their tasks, inputs, and results from general trends in \ac{XAI}. Linking these two explainability realms would not only result in a more standardized approach to explanations in \acp{RS} but also in a direct application of methods from \ac{XAI} to \acp{MRS}.

In this section, we review definitions and concepts of explainability in \acp{RS}. Subsequently, we compare and connect them with the ones of \ac{XAI}. Note that this is not a survey of \ac{XAI} or explainable \acp{RS} as other valuable resources exist on this matter \cite{DBLP:journals/ftir/ZhangC20,arrieta2020explainable,guidotti2018survey,nunes2017systematic, Survey2007, gilpin2018explaining}. % CITE: Survey2007, gilpin2018explaining

\subsection{Definitions and goals of explainability for MRS}

What does it mean to \textit{explain a recommendation}? Within the \ac{RS} field, Tintarev et al. 
\cite{tintarev2015explaining} addresses this question with \quotee{to make clear by giving a detailed description}, and Zhang et al.~
\cite{DBLP:journals/ftir/ZhangC20} with \quotee{an explainable recommendation aims to answer the question of why}.
We can thus discern a role of explanations as complementary information to the recommendation. But these definitions are limited; for instance, ensuring \textit{fair recommendations} involves tracing the "why" of a recommendation, but only regarding certain critical
aspects (\eg potential gender biases) and it does not tell how to act upon them. As we develop next, "complementary information" and "fair recommendations" shape two of the many facets of explainability.

Borrowing general ideas from recent harmonization efforts of \ac{XAI} terms, it is more convenient to distinguish between explanation objects and goals. In particular,     
\textit{explanations} designate the result of an explanation system, they form an \quotee{interface between the system to explain and a target audience} \cite{guidotti2018survey}.
%We will discuss the system part on which they are focused in Section~\ref{sec:model_data}, and the many forms they can take in Section~\ref{sec:explanation_type}.
Quite interchangeably with \textit{explainability}, we will use the term \textit{interpretability}, with a more passive characteristic: a system can be interpretable -- \eg decision trees are often interpretable, neural networks are not.
The opposite notion is often referred to as
\textit{blackboxness}. We stress that automatically concluding that trees and linear regressions are interpretable and that neural networks are not is questionable.\footnote{The existence of a general interpretability/accuracy trade-off seems a myth \cite{lipton2018mythos, rudin2019stop}, despite its persistent mentions in some \ac{XAI} papers.} As we will see next, this depends on a precise formulation of explanation tasks that do not admit one-size-fits-all rules.

% The notion of audience is essential
The previous mention of "audience" is essential, since a given explanation type may only convey meaningful information to specific people. In \ac{RS} research, the target audience of explanations is usually end-users as they are the targets of the recommendation decision and could be skeptical about it. %or doubtful of the results.
Nevertheless, other stakeholders may be interested in receiving explanations, \eg system designers and data scientists may inquire wherever their system bases its decisions on discriminatory biases from the data.

We shall continue with a cautionary tale: the disparate notions of explainability have led to many misuses of \ac{XAI} \cite{lipton2018mythos}. Because we usually do not have access to ground-truth explanations in the wild, and realistically will not in industrial contexts, many \ac{XAI} works have relied on intuitive notions of what their target explanations should be. This first makes evaluation difficult. As F. Doshi-Velez 
\cite{doshi2017towards} highlights, the relevance of explanations is often %merely 
suggested in a \quotee{you’ll know it when you see it} fashion, which paves the way to many \textit{confirmation biases}. Second, several counter-intuitive results have been unveiled. For instance, the widely agreed-upon idea that an interpretable model is more desirable than a blackbox one has been challenged: produced explanations -- similarly to model predictions -- may be misleading or biased \cite{rudin2019stop, dinu2020challenging, adebayo2018sanity, kaur2020interpreting, dombrowski2019explanations}. % CITE: dombrowski2019explanations
Moreover, without clear formulations of explanation tasks, how can several \ac{XAI} systems be compared? Can we actually quantify interpretability and explanation quality? Can the relevance of proposed interpretation metrics be assessed? How can we detect misinterpretations and explanations based on spurious mechanisms? All those questions circle back to the %very 
definition of explanations. 

\begin{figure}[h]
    \centering
    \includegraphics[width=0.85\linewidth]{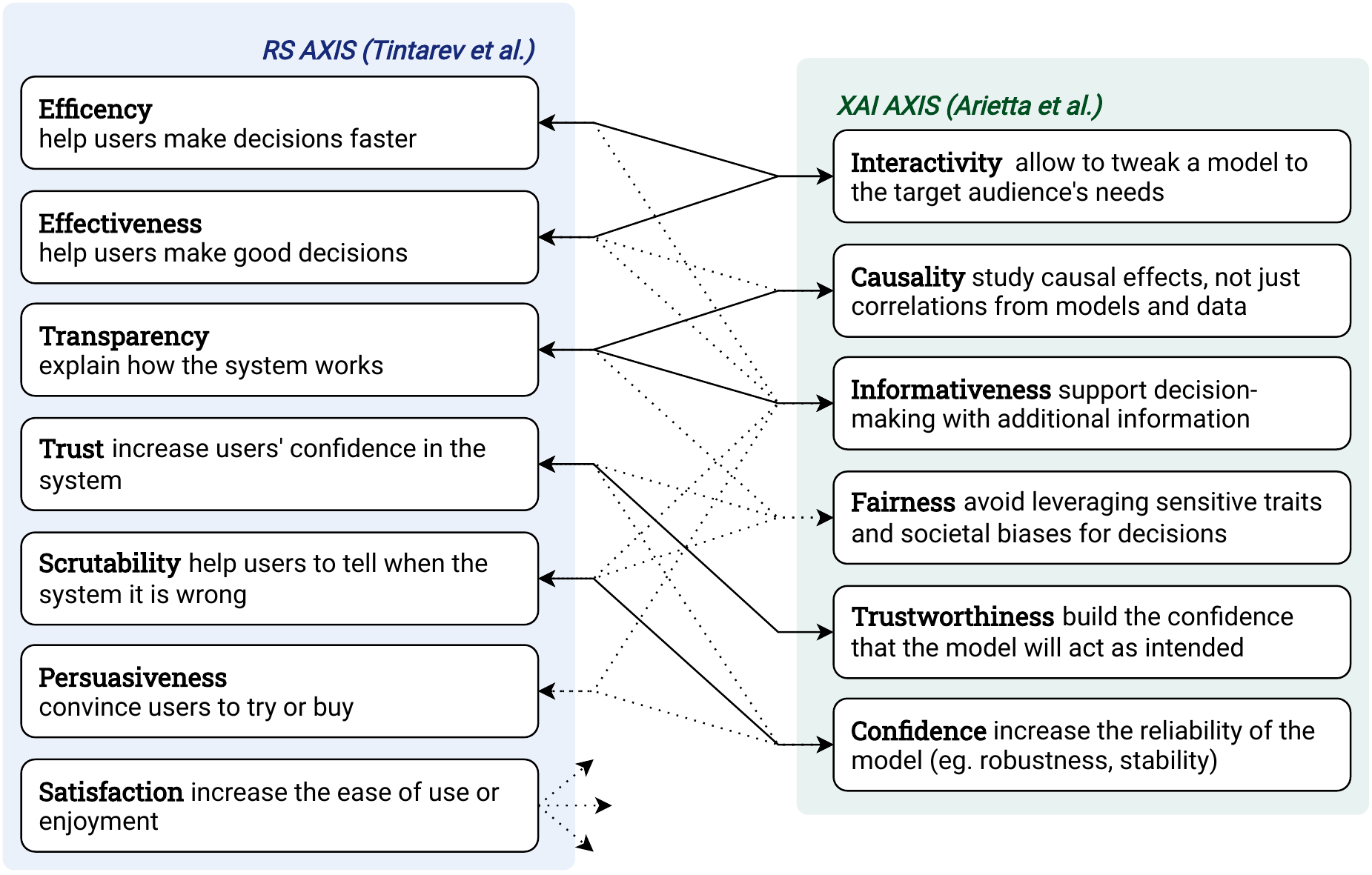}
    \caption{\textbf{RS and XAI explanation goals and linking}. We display "one-liner" definitions for conciseness. A solid line indicates a strong correspondence, a dotted line a weaker one that depends on the exact task and context. \textit{Satisfaction} could have been linked to everything but links are omitted for clarity.}
    \label{fig:links_axis}
\end{figure}

In addressing these questions, the concept of \textit{incompleteness} was proposed.
Its purpose is to characterize the "missing piece" justifying the use of an explanation system \cite{doshi2017towards}.
Here, the literature of explainable \acp{RS} and general \ac{XAI} diverges. A distinction of the goals of \ac{RS} explanations is proposed in~\cite{tintarev2015explaining}, which delineates seven of them.
We can enrich this discussion with goals identified in general \ac{XAI} by Arietta et al
~\cite{arrieta2020explainable}. Both sets of goals are displayed in Figure \ref{fig:links_axis} with short definitions. We find that neither of the two may solely account for all \ac{MRS} purposes:
%As alluded in the beginning of this section, 
Explainable \ac{RS} goals mostly fall into the \textit{informativeness} category.
This has a broader scope than \ac{RS} \textit{transparency} that feels too focused on the decomposition of models' inner mechanisms. Furthermore, \ac{RS} goals have been found to be intercorrelated \cite{balog2020measuring}, in particular, \textit{satisfaction} being arguably a desired byproduct of any explanation method. 
That said, \textit{persuasiveness} is a strong dimension of \acp{RS} that is absent from general \ac{XAI}~\cite{ehrlich2011taking}; 
when aiming at transparency, creating a persuasive system may appear contradictory.% it should be considered when aiming at transparency to avoid instead creating a persuasive system~\cite{ehrlich2011taking}. % CITE: gilpin2018explaining

%Meanwhile, \ac{XAI} goals are more orthogonal and more readily refer to practical use-cases.
% As much as we would like to further discuss the links between those two framing, this is impossible as potential links have to be particularized to each task and setting, which is not the purpose of our study.

\begin{figure}[h]
    \centering
    \includegraphics[width=\linewidth]{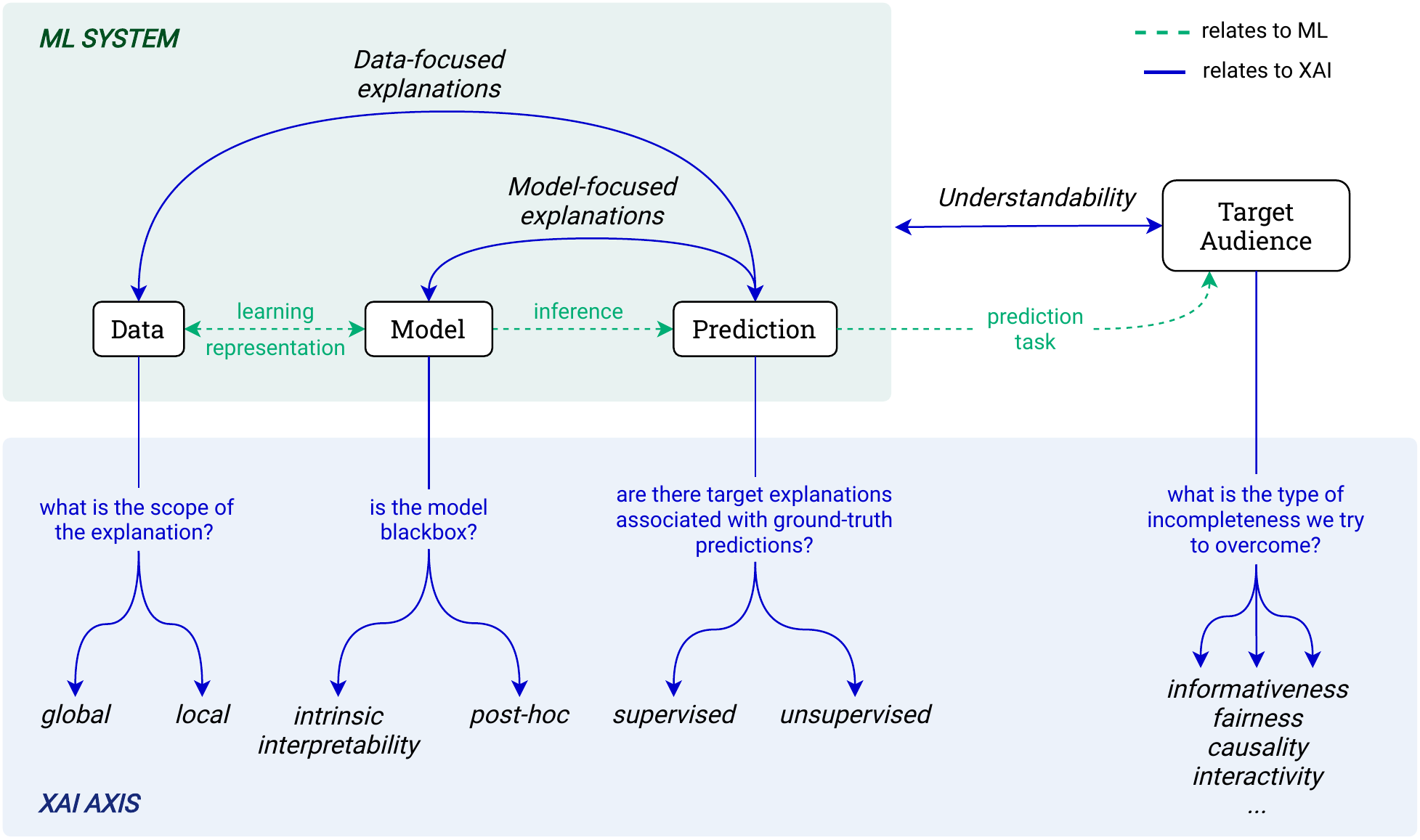}
    \caption{\textbf{Overview of \ac{XAI} notions. %discussed in Section~\ref{sec:goals}
    } In the upper part, a ML model is trained on data and used to make predictions %about it
    . Beyond prediction, if the model alone is insufficient \wrt an underlying human-grounded application, the use of an XAI method will be justified. The specification of the target audience delineates incompleteness to be addressed through explanations, along different explanation axis (lower part).}
    \label{fig:map_XAI}
\end{figure}

Identifying those goals is crucial because explanations may simply not be needed if incompleteness is not an issue.
Evaluation should then be conducted with regard to each targeted incompleteness, to avoid mismatched objectives.
Lastly, the concept of \textit{understandability} simply bridges the gap between a chosen \ac{XAI} system and this new notion of goal/incompleteness being addressed for a target audience. All these notions are illustrated and placed accordingly in Figure~\ref{fig:map_XAI}. We discuss additional taxonomic axes for \ac{XAI} in subsequent paragraphs. %Adaptations to the music field will be discussed in the next section.

As a final note, explainability can be framed through an interesting take from Michael Jordan on the future of machine learning.\footnote{Math \& IA seminar: \url{https://vimeo.com/522733917}} The goal of \ac{XAI} is not only for decision-makers to understand model predictions, but to allow a back and forth interaction between the two. \textit{Why} do you make this decision? \textit{What if} this aspect was different? \textit{Then} what if this aspect was different? \quotee{Consequential decision involves thinking about new facts that were never put in the original data, that are relevant to the current situation.} The purpose of the discussion on \textit{incompleteness} and \textit{understandability} is to go beyond the view of explainability as a mere complementary prediction, but to allow this reciprocal gain of knowledge between several actors.

\subsection{Local/Global scope}
\label{sec:local_global}
As \acp{MRS} commonly provide numerous recommendations, there is a focal distinction to make in the explanations scope: local \vs global \cite{doshi2017towards}.
%each of which represent a specific decision/prediction of the model. This leads to a focal distinction in \ac{RS} explainability based on the scope of the explanation, \textit{local} \vs \textit{global} \cite{doshi2017towards}.
\textit{Local} or instance-wise explanations target the decision of the model for a specific input-recommendation pair, \eg explaining that a track was recommended to an end-user because some of its features matched. Local explanations must be tailored to each individual prediction. 
This type of explanation is aligned with the European General Data Protection Regulation (GDPR) "Right to explanation" \cite{gdpr}, which entitles users to inquire about the reasoning behind the outcome of an algorithm, hence supporting informativeness as an explainability goal. 
%Furthermore, since the local framing acts on a user granularity, it also supports the interactivity goal. 

In contrast, \textit{global} explanations provide a big picture of the model logic, covering multiple model decisions.
For instance, estimating clusters of machine-learned user embeddings may help rationalize the behavior of the \ac{MRS} within several general communities. This broad view of the model is necessary to detect systematic biases of the model (addressing fairness goals) and %, overall,
to examine wherever a model is suitable for deployment (addressing informativeness, trustworthiness, and confidence). Lastly, note that the two types % of explanations 
may be linked: it is sometimes relevant to craft a global explanation by providing multiple %diverse 
local explanations \cite{ribeiro2016should}.

\subsection{Intrinsic/Post-hoc interpretability} \label{sec:intrinsic_posthoc}
We can also distinguish explanation systems \wrt whether interpretability should be an inherent part of the \ac{RS}--\textit{intrinsic} interpretability, or should be provided as an addition to an already working \ac{RS}--\textit{post-hoc} interpretability.

\textit{Intrinsic} interpretability refers to the ability of the \ac{RS} to provide sufficient information to make its inner functioning clear to a specific audience \cite{arrieta2020explainable}. In this case, the explanations coincide with the model. Being inherent in the model, intrinsic interpretability has to be planned in advance, making it a component of the model design. For instance, an Item-\textit{k-Nearest Neighbours} model recommends artists because they are similar to the ones the user listened to, thus allowing explanations such as \examp{We recommend you \texttt{<artist>} because it is similar to \texttt{<artist(s)>}}. 

\textit{Post-hoc} or extrinsic interpretability refers to the use of external \ac{XAI} to yield knowledge from a blackbox %\ac{RS} 
model. It can be considered as \textit{reverse engineering} the model \cite{guidotti2018survey}. For example, the recommendations of a blackbox model can be explained by making a post-hoc selection of the relevant features that lead to the recommendation; they offer explanations such as \examp{We recommend you this because it has \texttt{<feature(s)>} you may like}. Both intrinsic and post-hoc views are affiliated with the concept of transparency, thus supporting informativeness, causality, and confidence. However, post-hoc explanations are hampered by their externalness and require an additional check of their faithfulness to the studied model.
Yet, compared to intrinsic, they %considerably
disentangle model design from explanation design, allowing to consider XAI systems in a later stage, or to apply them to already working models.

\subsection{Un/supervised explanations}\label{sec:unsupervised_goal}
We often think of \ac{XAI} methods as being unsupervised. Particularly on the end-user side, it is arduous to guess which could be the \textit{ground-truth} explanations for the user since their judgment of what a good explanation is may be biased \cite{miller2019insights}. Nevertheless, target explanations are sometimes available \cite{balog2020measuring}. % cite: chen2019personalized
But far from making it a supervised task, 
our goal is not only to make explanation predictions but to address an incompleteness; the relevance of the target predictions thus has to be questioned. 
Do these really address our needs \wrt to incompleteness? Or are they a proxy for it? In the latter case, how do we assert/evaluate their \textit{understandability} \wrt our goal? We present two ideas from \ac{XAI} for supervised explanations in the image domain that could be applied to \ac{MRS}.

In the field of image classification, some datasets gather images with textual descriptions. 
%.. composed by set of words that can be matched against corresponding visual aspects.
%By leveraging them, 
Each set of words can be matched against corresponding visual aspects in the images, enabling to generate \textit{visual explanations} for class prediction of unseen instances %unseen instance class predictions 
through RNN-generated texts \cite{hendricks2016generating}. The explanations are evaluated against held-out test descriptions. Here, the concept of explanation is driven by two desiderata, first as a way to link different modalities of a same object -- image and text, and second as a rationale that conveys useful information by yielding class-specific information that differentiate it from other classes. Obtaining this informative \textit{discriminative} quality is tricky in an %(blind)
unsupervised setting. The multimodality of music data (\eg audio, lyrics, users, %communities, 
playlists) makes it a good candidate for this paradigm.

We can identify another line of supervised explanations as linking different conceptual levels. The \textit{TCAV} method \cite{kim2018interpretability}, for instance, allows to check predictions against human-understandable concepts, \eg how much the model prediction for an image of a zebra is sensitive to %concept of 
"stripeness". Again, there is an interesting link to music: there is a known and unresolved \textit{semantic gap} between low-level data (\ie audio signal) and its correspondence to high-level descriptions (\eg genre, mood) \cite{celma2006bridging}.

\subsection{Model/Data}
\label{sec:model_data}
We conclude this section with a paramount yet subtle distinction that is prone to be overlooked: are the explanations related to the \ac{RS} model processing or to the data it represents?
%More than splitting methods into two realms, it is interesting to inspect the links between the two philosophies.

\textit{Model explanations}, on one side, focus on a learned representation and parameters and aim at making sense out of it. With a mild exaggeration, to the question \examp{why is this track recommended by the \ac{MRS} given my history?} a model-focused answer of a %classical
\ac{RS} might be \examp{it maximizes the probability of being co-listened with your history, considering all other users listening history% track patterns
}. \textit{Data explanations}, on the other side, would rather focus on \examp{why are those items co-listened in the first place?}.
The trained model by itself is less interesting than the goal of uncovering "natural mechanism[s] in the world" \cite{chen2020true}. 
%aiming at uncovering "natural mechanism in the world" \cite{chen2020true}.
In practice, in the first case, the model inspection may expose irregularities and lead to adjust its architecture and regularization (\eg balancing fairness trade-off parameters); in the second case, the model plays the role of a proxy representation of data, detected errors would more suitably be attributed to a misrepresentation of input data (\eg feature engineering for a better matrix factorization), and the ultimate goal is to find a structure that is credible given prior knowledge of the problem.

These aspects are often entangled. Explaining the model provides little information with noisy data, and explaining the data may be misleading if the model assumptions do not capture salient aspects (\eg correlation instead of causation). It is a widespread fallacy to explain a model (which is often easier, particularly when using transparent models), when the true underlying objective is to explain data.
As a corollary, critics of \ac{XAI} often oscillate between "the method is unreasonable for explaining the model" (\eg randomizing the model's weights does not change the explanation \cite{adebayo2018sanity}) and "the produced explanations, though relevant for the model, do not make sense for humans", without explicitly mentioning this duality \cite{praher2021veracity}.

%Problems occur when those two paradigms overlap since they may lead to mismatched objectives.
%In essence, the way a model optimally reasons about data does not necessarily holds any truth about the way data is effectively structured. %For instance, decision trees are transparent: it is straightforward to provide a \textit{model}-explanation based on each logic branching (\eg it first found that $x_1 > 5$, then that $x_2 < 10$, thus it returned $3$). %But at the same time, trees notably suffer from identifiability issues. On rerun, different trees may be returned, changing the yielded explanations. This is evidence that doubts should be raised about their ability to extract general knowledge about data. %We can push the reasoning further by borrowing ideas from the causal field: trees do not account for redundant, confounded or mediated variables; they can exchangeably arrange causes and effects from root to leaves, and can link together coincidental effects. %It is a widespread fallacy to explain a model (which is often easier, particularly when using interpretable models), when the objective is to explain data. It is important to at least understand that this distinction exists to avoid falling into this trap.

\section{Making music recommender systems explainable}
\label{sec:explainabilityMRS}

In the previous section, we have drawn links between explainability in \acp{RS} and \ac{XAI}, and presented different definitions. Bearing these definitions in mind, we now study different ways \acp{MRS} can be made more explainable. We start with a general overview of possible explanation methods for \acp{MRS}, then discuss the adaptability of three relevant explanation paradigms to \acp{MRS}. %Inherent industrial challenges and other difficulties to effectively implement those methods to real -- usually complex and multi-component -- \acp{MRS} will be examined later.

\subsection{Overview of explanation methods for MRSs}
\label{sec:explanation_type}
We want to provide the reader with a short background on existing explanation methods for \acp{RS}, and then discuss how the latter are particularized for \ac{MRS}. 
%Specifically, we present a variety of explainability ideas from this literature through the pragmatic lens of the form they can take.

\subsubsection*{Explanations of \acp{RS}}

%Explanations provided with recommendations come in different styles and informed by various sources. On this matter,
Zhang and Chen 
\cite{DBLP:journals/ftir/ZhangC20} characterizes six \acp{RS} explanation types. First, \textit{relevant item or user} explanations, also called example-based explanations, are bonded with item-based or user-based collaborative filtering \cite{behnoush2018using}. % CITE: behnoush2018using
Thus, a recommendation is motivated either by the similarity of the item to other items previously liked by the user, or by the affinity that similar users have towards the recommended item. 

Second, there are \textit{feature-based} explanations, which %contrary to the collaborative approaches, 
are associated with content-based recommendation algorithms. % CITE: hou2019explainable
Explanations are commonly shown as tags relevant to a user or an item~\cite{DBLP:journals/ftir/ZhangC20}. 
\textit{Opinion-based} explanations focus on relevant aspects of the recommended item~\cite{zhang2014do,wang2018explainable}, which can be enriched with a sentiment~\cite{zhang2014do}. 
In contrast to feature-based explanations leveraging item metadata or user profiles, opinion-based aspects are mined from reviews or social media posts.
%Additionally, each aspect can be associated with an opinion, which comes in the form of a sentiment regarding a specific characteristic (\eg screen-clear-positive) \cite{zhang2014do}.
 
Further, we also distinguish sentence, visual, and social explanations. 
\textit{Sentence} explanations can be predefined templates with placeholders regarding features or aspects/opinions filled on-the-fly depending on the recommendation or specific user (\eg \examp{We recommend this item because its \texttt{[good/excellent]} \texttt{[feature]} matches with your \texttt{[emphasize/taste]} on \texttt{[feature]}}) \cite{wang2018explainable}. Alternatively, sentence explanations can be generated from scratch using language models trained on reviews \cite{costa2018automatic}. 
\textit{Visual} explanations appear as images or visual elements often accompanied by text \cite{chen2019personalized,andjelkovic2016moodplay}. 
Image regions or caption words that explain the recommendation could be highlighted \cite{chen2019personalized}. 
\textit{Social} explanations mention either the user's friends who liked the recommended item \cite{sharma2013do} or their overall number.

\subsubsection*{Extension to \acp{MRS}}
% With voice assistants becoming increasingly popular, researchers are investigating audio-based explanations in \acp{MRS}.
% A first line of work, proposes \textit{listenable explanations} \cite{behrooz2019augmenting}, inspired from radio shows in which hosts provide information about played tracks for creating transitions.
% Alternatively, item parts such as track snippets focusing on a particular audio source (\eg instrument or voice \cite{melchiorre2021lemons}) can be emphasized as reasons for recommendation.   
Explanations in \acp{MRS} have multiple specificities.
First, they can be based on \textit{audio}. As voice assistants are becoming increasingly popular in music consumption \cite{sciuto2018hey}, researchers have been looking into how to augment recommendations with audio music explanations. One line of work proposes \textit{listenable explanations} \cite{behrooz2019augmenting}, inspired from radio shows in which hosts provide information about played tracks for creating transitions. Alternatively, item parts such as track snippets focusing on a particular audio source (\eg instrument or voice \cite{melchiorre2021lemons}) can be emphasized as reasons for recommendation.

Second, whenever recommendations are provided as collections of items (\eg playlists), explanation generation can be modeled as playlist captioning (\ie the automatic generation of a title and/or a description of the playlist) \cite{choi2020prediction} or playlist stories generation \cite{behrooz2019augmenting}. Existing work usually relies on predefined textual templates~\cite{behrooz2019augmenting}.

Third, music explanations are rarely informed by a unique data source. \acp{KG} are constructed from external sources and used for explanations \cite{oramas2016information}. Information sources leveraged in existing work are: user-generated text such as music descriptions \cite{zhao2019personalized}, existing knowledge bases like MusicBrainz or Wikipedia, \cite{moon2019opendialkg}, tags describing items or users \cite{zhao2019personalized,kouki2019personalized}, social information such as users' friends \cite{kouki2019personalized,sharma2013do}, audio features \cite{andjelkovic2016moodplay}, or pre-trained tag embeddings \cite{andjelkovic2016moodplay}.

\begin{figure}[h]
    \centering
    \includegraphics[width=\linewidth]{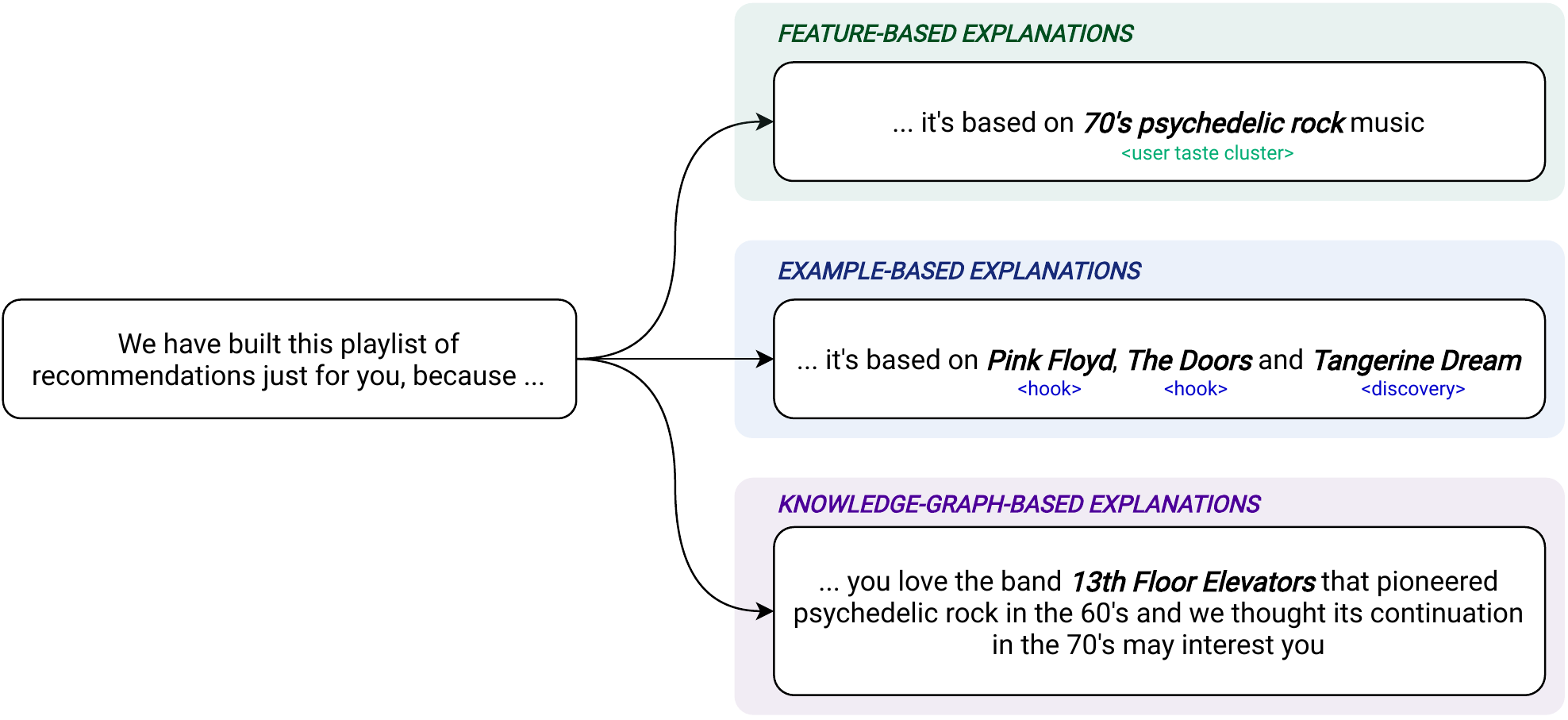}
    \caption{\textbf{Summary of explanation types.} We show some \textit{informativeness}-oriented explanation examples that may be provided to an \textit{end-user} for which a personalized playlist has been generated.}
    \label{fig:map_sec3}
\end{figure}

We next discuss in detail feature-based explanations (Section~\ref{sec:feature_based_explanation}), example-based explanations (Section~\ref{sec:example_based_prototype}), and graph-based explanations (Section~\ref{sec:graphbased_expl}). We refer to Figure \ref{fig:map_sec3} for examples of each explanation type. 

\subsection{Feature-based explanations}\label{sec:feature_based_explanation}

\acp{MRS} rely on multi-modal information (or features) in order to provide personalized recommendations to users. It is therefore legitimate to ask \textit{which features are most responsible for the generated recommendation}. Feature-based explanations aim at answering this question by identifying a minimal subset of features that are relevant for the recommendation. For instance, such explanations may be \examp{We recommend you this song because it is '90s rock, a combo of era and genre you enjoy listening to.} where the \textit{genre} and \textit{era} represent the relevant features.

\subsubsection*{Relevance}
Feature-based explanations are only relevant if the features are themselves interpretable. Furthermore, feature selection is an NP-hard problem \cite{natarajan1995sparse} and real-word applications necessarily rely on feature assumptions: \eg a limited number of interacting features \multipleoptcite{chen2018learning}{lou2013accurate}, group or structure coherence \multipleoptcite{afchar2020making}{zhao2009composite}, % CITE: huang2010benefit,
feature independence \cite{ribeiro2016should}, or first order approximations \cite{simonyan2013deep}.

\subsubsection*{Applications}

Frequently, considered features are selected and ranked through a relevance score. More than just the top-contributing features, displaying or visualizing all the scores is a common practice among data scientists, acting as an encompassing explanation \cite{kaur2020interpreting}, though the information overload may be misleading \cite{poursabzi2018manipulating}.
Note that "relevance" for a feature is a polysemous term that inherently depends on the used selection method. As illustration, both SHAP~\cite{lundberg2017unified} and L2X~\cite{chen2018learning} assign relevance scores to single features, however, while SHAP expresses relevance in terms of marginalized contributions of features across all possible subsets, L2X encodes relevance as a notion of informativeness on the response variable through maximizing mutual information. We refer to \cite{blum1997selection, covert2020feature, sundararajan2020many} % CITE: blum1997selection,
for surveys and further details on selection methods.

Applied to \acp{MRS}, feature explanations may be related to users, items, to the context, or a combination of the previous.
\textit{User features} stretch from reasonably static characteristics (\eg country of origin, age group, personality) to constantly-changing traits (\eg tastes, recent interests, mood). These features offer a fertile ground for tailored recommendations, and thus tailored explanations such as \examp{We recommend you this track because it suits your current emotional state} or \examp{... because of your country of origin}. However, the effectiveness of these explanations may be hindered by unreliable estimates of some user's variables, notably dynamic ones.
Fairness-wise, societal biases in \acp{RS} often stem from the usage of sensitive user features, an analysis of their impact on recommendations being crucial to be able to temper with them. 

Alternatively, \textit{item features} are usually more objective and come in different types and granularity. For example, audio features cover low-level features (\eg spectrograms, beats) \cite{muller2015fundamentals, pons2017end}, % CITE: pons2017end
mid-level features (\eg articulation, melodiousness) \cite{aljanaki2018data, patwari2020semantically}, % CITE: patwari2020semantically
or high-level features (\eg danceability, emotion) \cite{kim2020butter, melchiorre2020personality}, % CITE: patwari2020semantically
as well as metadata (\eg genre, social tags) \cite{knees2016music}. % CITE: knees2013survey
Explanations involving these features are strongly tied to content-based recommendation \cite{DBLP:journals/ftir/ZhangC20} as they directly match the user's preference profile (\eg \examp{We recommend you this because it has the tempo/genre you like}). %For active users, allowing to tweak the weights of different \ac{MRS} item aspects may support interactivity (\eg giving importance to the mood but less to proposed music genre continuity).

Lastly, miscellaneous \textit{context features} may be suited for generating personalized explanations. Time and location, being the most popular ones, provide a sound contextualization for the recommendation such as \examp{This techno masterpiece is perfect for tonight's Friday's party!} or \examp{Since you are doing home-office these days, we recommend you this 'Work from Home' playlist}.

\subsubsection*{Evaluation}

Feature-based explanations are tied to a chosen definition of relevance. Their approximations can be compared to full computation results, when affordable.
However, it may be tricky to evaluate whether the relevance scores themselves translate to true relevance. What is relevant for a trained model indeed reveals correlated events in the data, with the risk of returning spurious relations instead of causal truth about the data. 
As another pitfall, many feature selection methods do not handle intercorrelated features well \cite{yu2003feature}, which are however common with \acp{MRS}.

\subsection{Example-based explanations}\label{sec:example_based_prototype}
%Example-based explanations for \acp{RS} are sets of instances that provide rationales for recommendations. https://www.overleaf.com/project/5ff58a93f1c9452fbb4df255
In \acp{MRS}, example-based explanations are a very common type of explanation, that can be reduced to the use of the sentence template \examp{We recommend you \texttt{<this new item>} because of \texttt{<its similarity>} to \texttt{<meaningful item(s)>}}. They are conceptually tied to Case-based \ac{RS} \optcite{bridge2005case}.

\subsubsection*{Relevance}
Similarly to feature-based explanations, they are only relevant if the given examples are themselves interpretable for the target audience. This includes returning items that are known to the user: \eg from a set of liked or previously interacted items, or from broadly-known items.% -- \eg naming \examp{Mozart} will ring a bell to everyone.

\subsubsection*{Applications}
Regarding \textit{example types}, it is common to see \textit{artist} examples as they convey a general sense of genre, temporal period, or style. 
Relevant-user examples were popular in the past decades as they have the interesting social twist of fostering users' curiosity to find pairs with similar tastes.
They have gradually vanished from most music and video streaming platforms since they were found less convincing and accurate than item-based explanations, and in turn may have a negative impact on \textit{trustworthiness} \cite{herlocker2000explaining}. Nevertheless, limiting social explanations to close circles was found more relevant (\eg \examp{recommended tracks recently discovered by your friends}).
Other than textual modalities, explanations in \acp{MRS} include displaying album covers, which may convey information about the style or even allow to recognize record labels (\eg \textit{Deutsche Grammophon}, \textit{Blue Note}). Short audio thumbnails are also a promising way to provide explanations that cannot be otherwise expressed with words \cite{melchiorre2021lemons}.

As for \textit{similarity relations}, we note they may not be explicitly stated in the explanation, or in some cases cannot even be stated. This is particularly true for \acp{RS} basing their recommendation on co-listening data. With the same causality counterpoints as before; a co-listening may be coincidental, confounded by external factors, or, more pragmatically, may result from noisy metadata and inattentive users.
% may result from a user with very eclectic tastes that is not necessarily relevant for another one.
With deep learning models that compute non-linear similarity metrics (\eg the NeuCF method \cite{he2017neural}),
it gets trickier as we are faced with an added blackboxness issue.

This can lead to explanation examples that feel cryptic to the user. Recent works in \ac{KG}-based recommendations are a way to alleviate this issue, we will discuss them in Section~\ref{sec:graphbased_expl}. Another lead lies in the disentanglement of the embeddings' latent dimensions that help rationalize proximity according to explicit concepts (\eg audio features, genre, instrumentation) \cite{lee2020disentangled}. Attention-based mechanisms are also a promising way of providing recommendations based on the selection of a reasonably small and contextual subset of neighbors \cite{kang2018self}, though claims on the interpretability of attention are disputed \cite{serrano2019attention}.

\subsubsection*{Evaluation}
It is useful to evaluate the \textit{discriminativeness} of examples. Indeed, example-based explanations are affected by popularity biases, which hampers \textit{informativeness}. As illustration, \examp{The Beatles} are streamed by many users with diverse profiles, thus appearing in many co-listening relations and are likely to emerge as similar neighbors. But using them as examples consequently lacks representativeness. On the other end, examples of niche artists are double-edged: they can yield a powerful feeling of understanding of user taste, but if the recommended item falls too far off the example style, and as the user sensibly connects with them, the explanation can feel disappointing. %Good examples fall in-between.

Then, one should distinguish items coming from an explicit elicitation (\eg liked artists) and implicit preferences. The former are often meaningful to users but may make them feel trapped in a recommendation bubble, while the latter are more diverse but potentially lack direct connection to users, affecting trust in explanations.

Examples can also be useful for persuasiveness goals. It may be interesting, for instance, to provide a set of examples that include an item that is well-known to the user (acting as a \textit{hook}) and unknown or weakly interacted ones (acting as discoveries). This principle is quite common in radio "clock" programming, where alternating \textit{power songs} and discoveries has been shown to be a powerful tool to keep users engaged.

%First, we have hinted that relevance is partly due to \textbf{usage and habits}. It is common for MRS that explanations carry information about either the \textit{genre}, \textit{era} or \textit{style} of recommended items. It would feel strange if we saw a generated playlist based on the set of artists "\textit{Sex Pistols}, \textit{Mozart}, \textit{Janis Joplin} and \textit{Lili Boulanger}", as the link between those artists does not seem to fall into these three prior templated similarity relations we are used to handle. Namely here, that those artists all died pretty young and that their compositions are tainted with a looming -- almost foreshadowed -- presence of death. Crafting new similarity relations goes hand in hand with a good UX to feel relevant to users, and pedagogy if the goal is to push users to \textbf{discoveries}.

\subsection{Graph-based explanations}
\label{sec:graphbased_expl}

% Conceptually, the matching of users and items in a RS can also be modeled by graph-based approaches on bipartite graphs, with users on one side and items on the other. Thus, the RS task can be framed as....

Canonically, \acp{RS} match
users and items. It is therefore not surprising that %links are drawn to 
graph-based approaches on %for this setting is very reminiscent of a 
\textit{bipartite graphs} can be used, with users on one side and items on the other.
The recommendation task may indeed be framed as \textit{link prediction}: given links of interactions between users and items, which unseen links are then probable? Likewise, similar item recommendation can be formulated as the task of finding probable nearest neighbors %-- to possibly multiple degrees --
in a graph of users. %neighbors.

\subsubsection*{Relevance}

This framing can seem cumbersome, with a first strong hindrance that graph methods quickly get computationally expensive -- though some works have demonstrated industrial-scale applicability \cite{ying2018graph}. Second, notions of repeated music consumption, preference decay and accounting for a temporal dimension for sequential recommendations are tricky to incorporate into graphs.
Nevertheless, graph-methods possess an outstanding expressive power, especially for multi-relational data, enabling abundant new \ac{RS} applications.

Further justification lies in the {graph structure naturally found} in \ac{MRS} data. \textit{Vertically}, there is a natural hierarchy for musical items: tracks are organized into albums, that are themselves children of artists, that can be regrouped into genre, style, and time period, or any complex multi-leveled music ontology. Users exhibit a similar hierarchy, we can often assign them to several clusters of interest, that are themselves linked to a given culture, country, or age category. \textit{Horizontally}, music item clusters act as islands of connected components, with central nodes being representative of a given style and having influence on surrounding artists. Weakly connected nodes denote niche artists, and nodes in-between clusters fuse several influences. The same reasoning may apply to users' communities and hierarchies.

\subsubsection*{Applications}
Graph analysis tools can be used to analyze node and edge structure. 
Detecting \textit{cliques} and using \textit{$k$-degeneracy} can help represent communities. \textit{Tripartite} and generally $n$-partite formulations allow to generalize canonical recommendation by handling more actors than items and users, for example considering artists and context. % to have multiway relations. 
\textit{Directional edges} can be leveraged to create graphs with asymmetrical relations and avoid recommending niche artists as being similar to very popular ones \cite{salha2019gravity}. Graph-specific \textit{embedding} techniques may be applied, \eg using random walks to train embeddings on more diverse sequences than observed data \cite{grover2016node2vec}. Other approaches are promising, such as analysis of graph structure for \textit{domain-transfer}; application of the \textit{traveling salesman problem} to find fluid playlist tracks orderings; or \textit{continual learning} by framing the addition of new users and items as new nodes that should not perturb far away regions of the graph. All these tools address the transparency issue, leading to more interpretable models.

Interpreting structure is mostly useful for an audience of researchers, % to better interpret and handle data
but recent advances in the field of \textit{knowledge-graph-based recommendations} show additional promising applications for end-users.
The term \ac{KG} refers to the use of external expert-knowledge to better understand the entities at hand within a \ac{RS} task and how they relate to one another \cite{dong2014knowledge}.
In the context of \acp{MRS}, available knowledge may include intra-music relations (\eg \examp{is sung by}, \examp{has music label}, \examp{belongs to genre}), and collaborative information (\eg \examp{often streamed with}, \examp{user taste belongs to cluster}).
The \ac{KG} can be applied to enhance the representation of items before recommendation. For instance, the latent space that is usually learned to compactly represent items can be structured to align with each item relations of the graph \cite{bordes2013translating, wang2014knowledge}. % CITE: wang2014knowledge
%For instance, if item $i$ is associated to $j$ through a given relation $l$ (\eg $i =$ \textit{"Meddle"}, $j =$ \textit{"Pink Floyd"}, $l = $ \textit{"is album of"}), denoting by $e_i, e_j, e_l$ their respective embedding vectors, we ensure that $e_i + e_l \simeq e_j$, which yields an embedding space with interpretable structure and grammar [transE, transH, transR].
However, \acp{RS} may still fail to leverage the full power of \acp{KG}, solely relying on enhanced representations. Instead, another approach is to directly incorporate \acp{KG} into the recommendation computation, which allows multi-hop reasoning.
For that, all paths (with a fixed maximum length) between a pair of user and item can be extracted, and their relevance estimated \cite{wang2019explainable}. % CITE: wang2019kgat
This enables to produce explanations corresponding to paths of high probability (\eg a path $\textrm{User}_i \xrightarrow[\textrm{listened to}]{} A \xrightarrow[\textrm{sung by}]{} \textrm{artist}_A \xrightarrow[\textrm{belongs to}]{} \textrm{label}_L \xleftarrow[\textrm{belongs to}]{} \textrm{artist}_B \xleftarrow[\textrm{sung by}]{} B$ translates for user $i$ as \textit{"Track B is recommended to you because it's similar to A you listened to before, which is sung by an artist belonging to the same indie music label $L$ as B."}).
For a complete survey of \ac{KG} methods, we refer to \cite{ji2021survey}.

\subsubsection*{Evaluation}
The efficiency of those techniques is conditioned on a good modeling of the involved entities (\ie nodes and links), deep knowledge engineering, and an accurate estimation of the paths' relevance while ensuring their interpretability. As a counter-example, a generic \examp{similar to} relation in a \ac{KG} does nothing for \textit{informativeness} as it is still a blackbox information, no matter the transparent relations before and after in the path.

Multi-hop reasoning that is permitted by graphs is a great opportunity to enhance discovery, which is known to impact \textit{effectiveness} and \textit{satisfaction} of \acp{RS} \cite{castells2011novelty}. But this requires crafting new metrics for relevance evaluation, which is still an open research topic~\cite{ge2010beyond}.

\acp{KG} are also a promising lead for \textit{causality}, as they can allow to model and estimate causal structures for data.

\subsection{Perspectives}

%To conclude this section, we provide a brief perspective on future approaches of explainability in \acp{MRS}.

Drawing inspiration from the recent success of GANs \cite{goodfellow2014generative}, we could consider \textit{generative explanations} in \acp{MRS}. In particular, assuming the audio content is available, a GAN-generated explanation may provide a listenable explanation of what the user tastes are like according to the model. Indeed, the explanation may be conditioned on some priors \cite{mirza2014conditional}, \eg what the user likes about metal or jazz, to provide reasonable explanations. However, these types of explanations are hampered by the demanding resources required to generate audios \cite{dhariwal2020jukebox}.

Another interesting direction is exploiting human concepts of musical understanding \cite{kim2018interpretability, chen2020concept}. % CITE: chen2020concept
For example, to understand how much the concept of 'rock' or 'happy' matters for the recommendation to a specific user. Beyond informativeness, this may also lead to uncovering bias in the datasets (\eg how much the concept of male artist matters for the recommendation).

%Some explanation systems were not applied to music, but we can give examples of how they may theoretically be. As a "perspective" section. Also, all previous sections were mostly informativeness-oriented explainability, which is indeed relevant as it is the most relevant for users and companies currently, but other needs are rising (eg. fairness):
%Methods: multimodal-recommenders (informativeness, example-based); embeddinsentanglement \cite{yin2020content,jianxin2019learning}, VAE, GANSpace (informativeness, high-level) ; fairness-inducing recommenders (fairness) ; debiased recommenders (trust? fairness?) ; conditional recommenders, sometimes called "style-based" (interactivity) ; domain-transfer (transferability).

Lastly, counterfactual or contrastive explanations not only pinpoint the causes of a model decision but also provide users with actionable levers to change the recommendation \cite{wachter2017counterfactual, miller2019insights, ustun2019actionable}. % CITE: wachter2017counterfactual,
Among the explanation types, counterfactual explanations may be considered best compliant to the GDPR \cite{gdpr} % CITE: wachter2017counterfactual
as they can provide a refined framework for fairness~\cite{kusner2017counterfactual}.

\section{Evaluating explanations}\label{sec:evaluation}

Evaluating \ac{MRS} explanations is paramount to assess whether the explanation goals (Section~\ref{sec:goals}) are met by the explanation methods (Section~\ref{sec:explainabilityMRS}).
This is an inherently hard task since it involves a multitude of factors, including the  targeted goals of the explanation (\ie asserting \textit{understandability}), the type of explanation (\eg whether the \ac{XAI} method works as intended), and the underlying \ac{RS} model (\eg checking whether we are trying to explain meaningful recommendations in the first place). 
We have discussed %touched upon 
some evaluation aspects in previous sections, specific to particular explanation dimensions and categories of methods.
While there exists no one-size-fits-all evaluation strategy, in the following, we provide some general guidelines, tailored to  %depending on 
the target audience of the explanation. %(end-users vs.~technical stakeholders). 
%In general, there is no one-size-fits-all evaluation, though we can provide some general guidelines depending on the target audience of the explanation. In particular, we first discuss some of the evaluation methods for the end-users, and second for internal \ac{MRS} stakeholders -- \eg data scientists. 

\subsection{Evaluating explanation from the end-user's perspective}\label{sec:end_user_eval}
Since \acp{RS} explanations mostly target end-consumers, it is legitimate to involve them in the evaluation procedure.
%Most \acp{RS} explanations target end-consumers. %, it is legitimate to involve them in the evaluation procedure. 
One straightforward way to evaluate such explanations is to conduct user studies \cite{knijnenburg2012explaining, maxwell2017designing} % CITE: maxwell2017designing
and assess if the explanations allow to address the targeted goals. We have argued in Section~\ref{sec:unsupervised_goal} that an explanation ground-truth is an evasive concept. %We now argue that 
Nevertheless, user studies can provide cues for what explanation types are best suited in specific domains, investigate research questions (\eg should we use explanations in visual or text form?), 
and can also detect practical misuses \cite{kaur2020interpreting}.

In the context of \acp{MRS}, %previous research based on 
user studies showed that visual explanations increase understandability \cite{andjelkovic2016moodplay} while social or sentence explanations are more persuasive \cite{sharma2013do}. However, providing too many details results in cognitive overload and is negatively perceived \cite{kouki2019personalized}. Also, persuasiveness does not necessarily correlate with the value recommendations have for the user. %how good a recommendation is for a user. 
For instance, a user following an artist recommendation because a friend likes it does not necessarily result in the user liking the artist. One suggestion to overcome this is by corroborating different types of explanations (\eg social with feature-based explanations)~\cite{sharma2013do}. % to increase the informativeness~\cite{sharma2013do}. 
Another solution is to enable conversations between user and system, so recommendations could be gradually improved with system's explanations and user's feedback \cite{zhao2019personalized}. % CITE: moon2019opendialkg

User studies in \acp{MRS} are typically either between-subject or within-subject. Studies of the first type split users in two  %or more 
groups: one does receive the explanation, the other does not \cite{millecamp2019explain}%cite: zhang2014explicit
. Hence, we can naturally quantify the effect of the explanation by comparing the results between groups. %This type of study is also inline with 
The prominent A/B testing frequently used in industry belongs to this study type, where a large basin of users is available and different %types of 
interfaces can be tested simultaneously. 
In contrast, %user studies based on 
within-subject experiments are used when only few users are available, especially outside the industry context. In these studies, each user is presented with all  explanation interfaces~\cite{oramas2016information,kouki2019personalized,millecamp2019explain,herlocker2000explaining, tsukuda2020explainable,vig2009tagsplanations,chang2016crowd}% CITE: tsukuda2020explainable,vig2009tagsplanations,chang2016crowd
, and one containing no explanation. %Though more convenient, 
%evaluation of these studies 
Such within-subject studies need to take care of possible confounding factors emerging from the subsequent interaction with different interfaces (\eg a user may feel lost interacting with a complex interface after seeing a very simple one). 

Another fundamental aspect of user studies are the type of measurements they employ \cite{knijnenburg2012explaining}, usually either behavioral, such as click-through-rates and time-spent-interacting \cite{zhao2019personalized,andjelkovic2016moodplay}, or attitudinal, for instance, surveys and semi-structured interviews \cite{kouki2019personalized,behrooz2019augmenting}. Generally, the measurement should be carefully tailored to the explanation goal(s). For example, if \textit{persuasiveness} and \textit{trustworthiness} are the most relevant explanation goals, we can assess the first via click-through-rate and the second through specific questions %in a questionnaire 
\eg \quotee{Do you trust the recommendation?}. In an industrial context, these measurements may be used as key performance indicators of the explanations, though little research has been carried out here beyond general users' satisfaction (\eg streaming time and weekly active users count). 

Lastly, music consumption is influenced by the user's personal characteristics and context (see Section~\ref{sec:MRS}), which also affect the reception of the explanations. It is, therefore, necessary to take them into account by ensuring a representative population sample. Research has considered different demographics (\eg gender, age group, and country)~\cite{behrooz2019augmenting,sharma2013do}, %cite tsukuda2020explainable
musical sophistication~\cite{millecamp2019explain}, listening habits~\cite{oramas2016information,andjelkovic2016moodplay}, and psychological traits such as personality \cite{kouki2019personalized} and need for cognition \cite{millecamp2019explain}.

%It is worth to mention that other types of end-user evaluations are also available, for example, conducting a pilot study with a restricted sample to gather preliminary results \cite{behrooz2019augmenting}.

\subsection{Evaluating explanation from the technical stakeholders' perspective}\label{sec:experts_eval}
%In this section we briefly mention some ideas regarding explanation evaluation applicable to a variety of explanation methods from the technical stakeholder's side of \ac{MRS} \eg engineers, analysts and regulators. 

Methods to evaluate explanations can also serve the technical stakeholder's side of \acp{MRS}, \eg engineers and data analysts. %, and regulators. 
%As we may have already insisted on, 
Technical -- offline -- evaluations, though more convenient to conduct than user studies, are prone %to lead 
to the adoption of sketchy intuitive metrics, %with risks of 
which can result in confirmation biases \cite{doshi2017towards, lipton2018mythos}.
Fortunately, some metrics for explainability are widely agreed upon and seldom %have not been often shown to 
lead to misinterpretations. For instance, the \textit{stability} of an explanation between re-estimations \cite{bansal2020sam}, its \textit{robustness} to small data changes \cite{kindermans2019reliability, alvarez2018robustness}, and its \textit{consistency} across several similar models \cite{fel2020representativity} appear to be reasonable minimal requirements for \ac{XAI}.
%-- except if the goal is to explicitly do otherwise (\eg \textit{serendipity} \cite{ge2010beyond}).
Similarly, \textit{sparsity} is often desirable for explanations since fewer parameters in the explanation translate to better cognitive handling~\cite{rudin2019stop}. %We have evoked 
\textit{Discriminativeness}  %, as another desiderata for explanations, 
is already a not-so-trivial requirement as some popular feature-based explanation methods were shown %have been flagged 
%not to change when inspecting
to result in the same explanations across several class predictions \cite{adebayo2018sanity}.
Other subtle sanity checks are necessary: %have been (and will be) revealed in other works, as 
\eg some ML models tend to leverage out-of-distribution artifacts and thus provide nonsensical explanations \cite{kumar2020problems}, which must be avoided.

In a semi-encouraging manner, some \ac{XAI} goals seem harder to achieve than to check. For instance, fairness objectives often stem from  %quantified and identified and 
measured biases (\eg disparity), the impact of which a fairness-inducing system can thus be quantified~\cite{frye2019asymmetric}. 
Note that this gets trickier for less tractable objectives (\eg minimizing environmental impact) or if a complete measurement is unavailable, costly, or requires time to witness a significant change. The same could be said for \textit{interactivity}, for instance by tracking the variety of tracks a user listens to after adopting the system.

Not every method can generate explanations for all items or users of a \acp{MRS}. Thus, it is useful to measure the \textit{coverage} of a method, \eg %For example, measuring 
how many explainable items are recommended in the top-k list for each user %through precision and recall metrics
\cite{abdollahi2016explainable, peake2018mining}. % CITE: peake2018mining
Likewise, computational efficiency of explanation generation should be taken into account \cite{chen2018learning}, particularly for time-sensitive use cases. % timely contexts of \acp{MRS}.

\section{Explainability challenges in an industrial context}\label{sec:challenges_industry}

In previous sections, we discussed different ways to make \acp{MRS} more explainable and to evaluate explanations. We now focus on the inherent challenges %and other difficulties
that arise in a real industrial context
when trying to implement these methods % to real \acp{MRS} in order
to explain recommendations to end-users.

\subsection{Explanations in real MRS}
\label{sec:explanations_real_MRS}
\begin{figure}
    \centering
    \includegraphics[width=14cm]{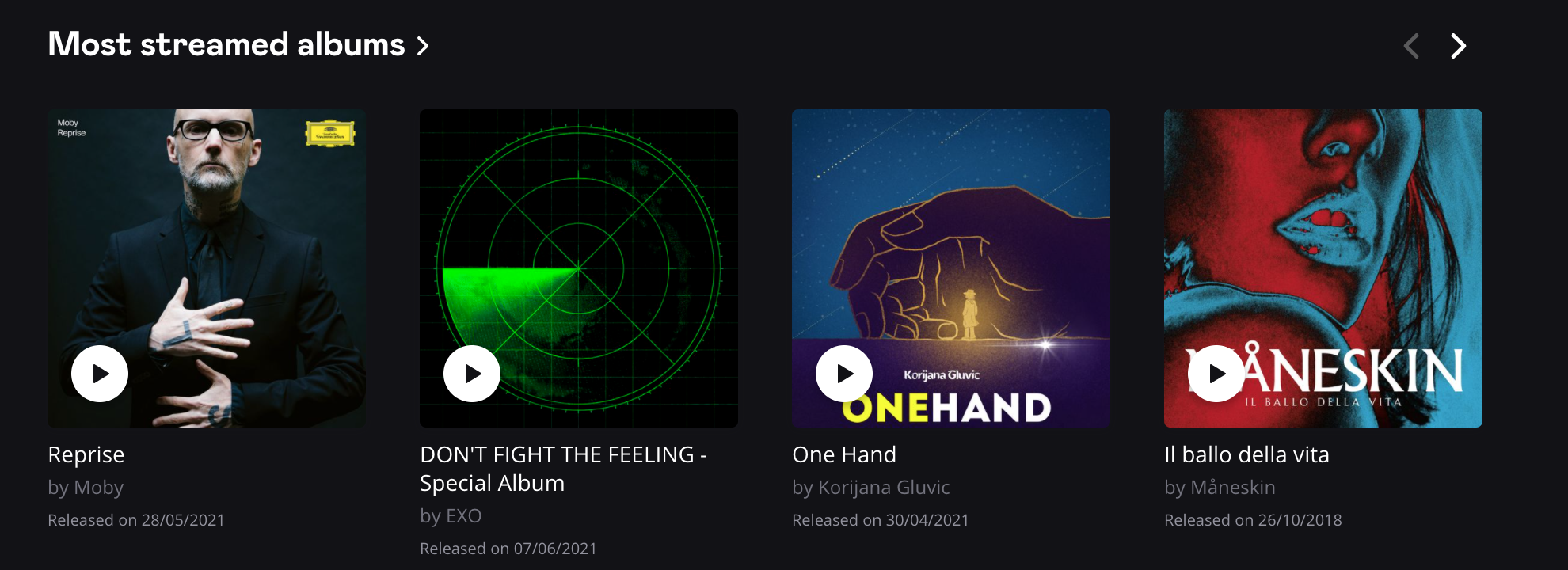}
    \includegraphics[width=14cm]{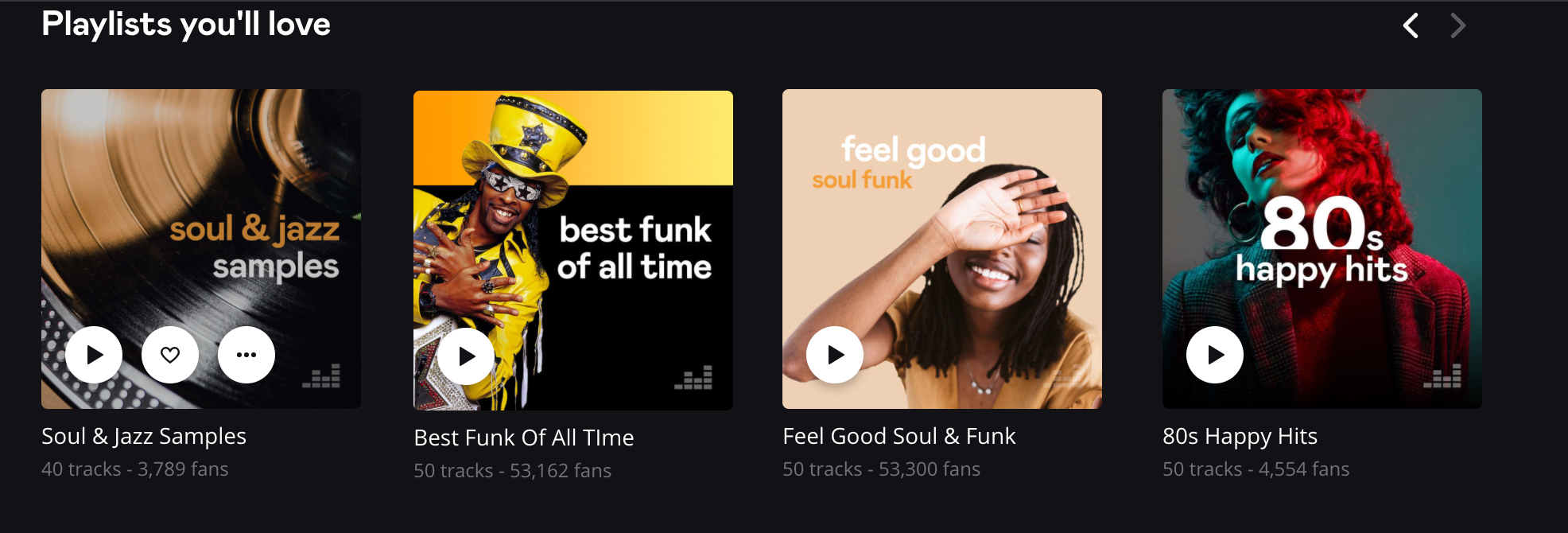}
    \includegraphics[width=14cm]{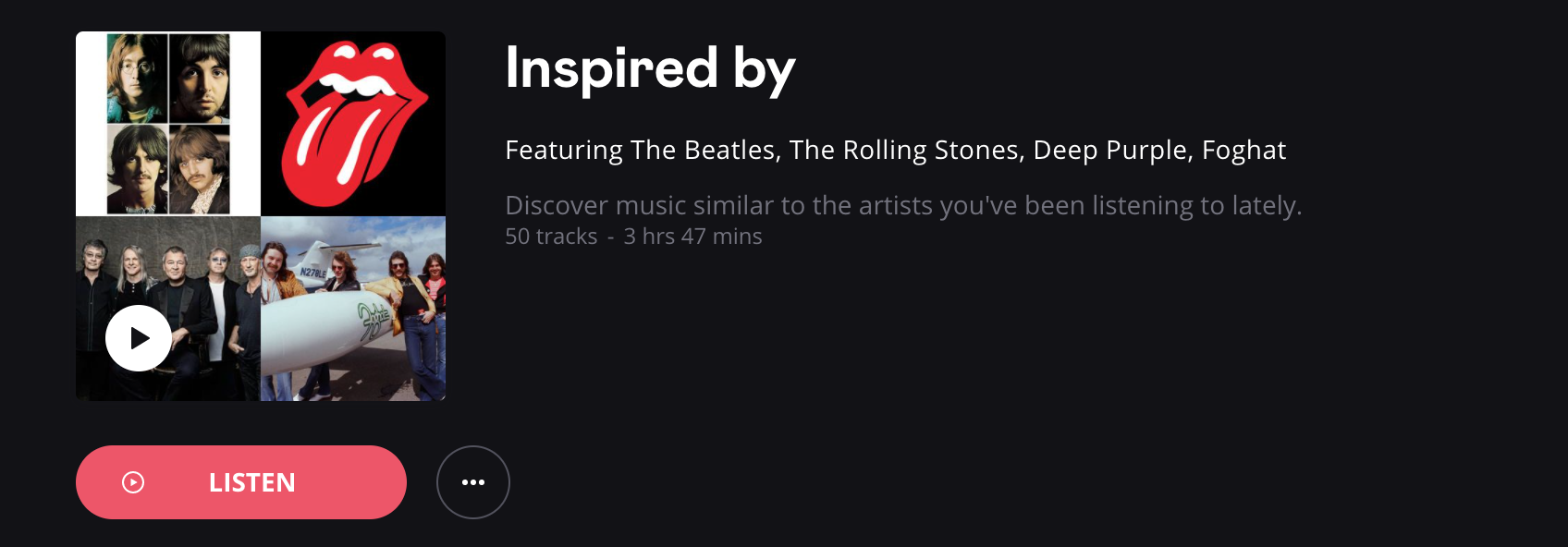}
    \caption{\textbf{Real-world recommendations with explanations.}}
    \label{fig:screenshot}
\end{figure}

Many providers of commercial music streaming services design their recommendation interface as \textit{swipeable carousels} \cite{bendada2020carousel}, namely sequences of sections that users can scroll. % for the user to scroll
%These sections titles can be seen as raw explanations of how the displayed content was picked for display. 
These carousels have titles that convey information to end-users such as:
\begin{itemize}
    \item Self-explanatory titles: \eg "Top 10", "Popular in your area", "Trending content" or "Recommended for you" that merely indicate the content selection process (Figure \ref{fig:screenshot} top).
    \item Feature-based explanations: \eg~"70's soul" or "Rock music" (Figure \ref{fig:screenshot} middle)
    \item Example-based explanations: \eg~"Because you like artist X",  "Because you listened to album Y" (Figure \ref{fig:screenshot} bottom)
\end{itemize}

%The first type is arguably self-explanatory and assumes no complex underlying recommendation algorithm at work. The second one is already more interesting, as the displayed category is informed by a user taste modeling process, it can be understood as a feature-based explanation (Section \ref{sec:feature_based_explanation}). However, this process is neither explained nor always explicitly mentioned. Finally the third category displays avatars of example-based explanations.

%Finally the last category implies that the content has been
%chosen considering 
%matched to 
%a user's preferences model and 
%this is indeed explicitly stated.
%is explicit about it.
%The "because you watched/listened to/liked" being a particular case where the justification is a content reference. This can be considered as a concise explanation targeted at end-user, serving a purpose of \textit{persuasiveness} \ref{sec:goals}.

Certainly, these simple and crude explanations are in contrast with the advanced explanation capabilities we have presented earlier. % in this paper.
Graph-based explanations, for instance, do not easily fit the headline formatting constraint, due to their length and complexity.
Therefore, they are quite uncommon in industrial systems though they represent a promising aspect of conversational \acp{MRS}. In the following, we further analyze this discrepancy between the scientific state of the art and the industrial realm.

\subsection{Overview of an industrial MRS} \label{sec:industrialMRS}
\begin{figure}
    \centering
    \includegraphics[width=14cm]{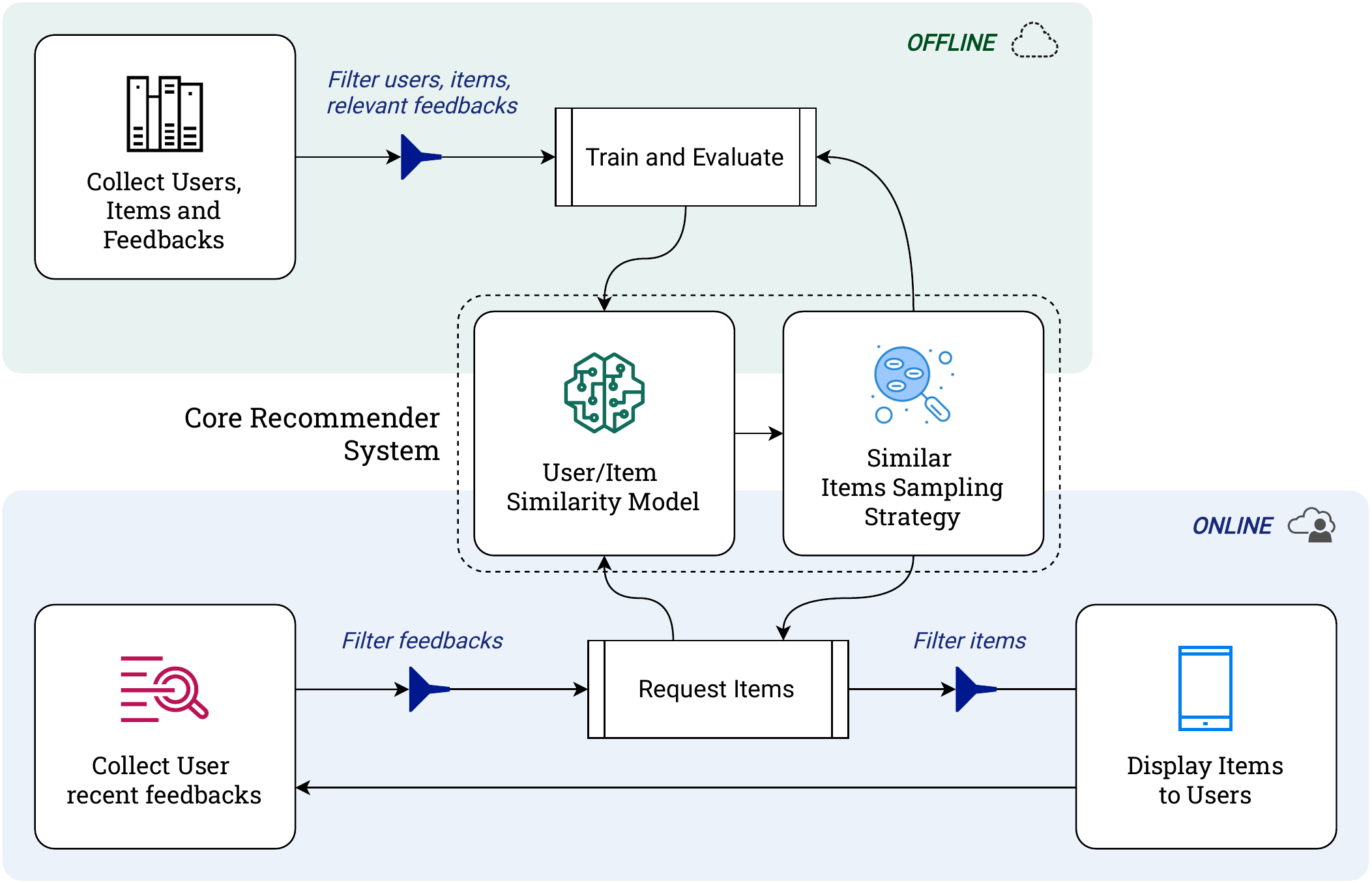}
    \caption{\textbf{Overview of an industrial \ac{MRS}.} Explaining the recommendation may require more than an explainable core \ac{MRS}.}
    \label{fig:industry_MRS}
\end{figure}

A simplistic view of an industrial \ac{MRS} is given in Figure~\ref{fig:industry_MRS}. Central to it is the \textit{Core Recommendation Engine} that models users and items affinities. Usually trained offline on a vast amount of user-item interactions, the system is then used online to generate item recommendations for each user accessing the service. This core \ac{MRS} is complemented by % modules engineered for 
heuristic filters and pre/post processing.

%When deployed in industrial setups, these systems are complemented by other modules engineered for heuristic filtering and pre/post processing.

To train and query the Core module, only a fraction of all available information about items and users %and their interactions
will be eventually used. For instance, users' metadata such as location, context or declared age can be used as-is, transformed (\eg quantized into broad areas or age buckets) or discarded. Items' data can be even more heavily processed. The audio signal can be subsampled, compressed, bounded, or normalized. Contextual information about the device, time, and location may be collected or inferred. Additionally, some systems leverage continuous user feedback in a session for online adaptation.
%Arguably the most impactful parameter choices are the ones controlling the sampling of user/item interactions. In particular the time-range of considered interactions and the filtering out of outliers or noisy points

Symmetrically, the direct output of the core \ac{RS} is not what the final user will be confronted with. Heuristics may be added, for instance to remove items that were already presented recently. In some contexts, enforcing contractual or legal obligations (such as the Digital Millennium Copyright Act rules for internet broadcasters \cite{1998DMCARules}) can also be necessary. Finally, product constraints in terms of display space on the device, connectivity status, or content availability issues can impact recommendations. % also play a role.

\begin{table}[h]
\emph{
``We recommended the song A by artist B to you because:
\begin{enumerate}[label=\textnormal{(\arabic*)}]
\item We considered your recent history (\eg 3 months) and that older interaction may no longer be relevant. Also, considering a longer time period would have been too computationally costly. \label{itm:history}
\item We also considered the recent history of many more users (not all of them, some were excluded because, for instance they had too few interactions, or peculiar activity patterns) to learn a representation space encoding similarity between artists with a machine learning system. \label{itm:similar}
\item The machine learning system learned to give close representations to artists that are co-listened by the same set of users and distant representations to artists that are not.\label{itm:embeddings}
\item We saw that you listened to songs by artists similar to artist B and you did not skip them which we interpreted as positive feedback. \label{itm:positive_feedback}
\item Eventually, you also explicitly liked artists similar to B or songs similar to A. \label{itm:like}
\item Some other songs that could have been very relevant in this context were discarded because you skipped them in a previous session. \label{itm:negative_feedback}
\item We sampled items in our representation space that are close to items to which you gave positive feedback and far from those with negative feedback. \label{itm:sampling}
%\item We selected this particular song of artist B because it has the same musical genre as the song you did not skip \label{itm:content}.
\item Song A and artist B also passed other heuristic filters (\eg regarding redundancy of recommended content, or a user personal blacklist).'' \label{itm:post-filter}
\end{enumerate}
}
\caption{\textbf{Honest recommendations explanation.}}
\label{table:transparent_explanation}
\end{table}

\subsection{Issues with explainability in industrial MRS}

If we were to provide a detailed description of the internals of an \ac{MRS}, destined to \textit{end-users} and using \textit{natural language}, it would probably look like the explanations provided in Table \ref{table:transparent_explanation}. While these may seem too detailed and almost provocative, they highlight a set of issues that we may face when trying to include explanations in an industrial \ac{MRS}. %and we will develop further starting from these examples.

%First let us remind that the theory of cognitive load applied to user interface design \cite{zhou2013affective} suggests that unnecessary display of information, going against ergonomics principle\cite{scapin1997ergonomic}, can be detrimental to the user's satisfaction. This \emph{less is more} design pattern is indeed widely adopted as a general good practice among industrial actors. In this regard, the display of explanations should be avoided or limited to a minimal and simple form to prevent cognitive overload.
%This need for design simplicity and conciseness is a very strong constraint for the integration of explainability in industrial \acp{MRS}.
%Simplicity can be a quite hard constraint in product design and limiting the number of interactions can be a feature thus limiting room for explainability.

\subsubsection*{Issues with engineering assumptions and design choices}

\acp{MRS} largely rely on implicit feedback and the engineering assumptions that come with their processing.
%For instance, most music services feature a continuous personalized radio mix. The \ac{MRS} 
For instance, most music services collect user feedbacks through basic interactions, namely \emph{skips}, \emph{likes}, \emph{dislikes}, listening history and navigation outside what was provided by the \ac{MRS} (such as music that was retrieved through the search engine). While \emph{dislikes} are rather self-explanatory, the intention of the user \emph{liking} a recommended item may not be that clear as users may use it for bookmarking songs. The intention behind a \emph{skip} is even more difficult to understand \cite{afchar2020making}, while \emph{skips} remain the most basic and common interactions. Thus, \ac{MRS} designers usually want to take advantage of them and enforce heuristic rules, \eg negatively weighting \emph{skips} and positively considering full-songs listenings (even though the music may have been played without someone actually listening to it).
%will usually make an assumption on top of these interactions, most of the time considering a \emph{skip} as a somewhat negative signal, such as explained in part \ref{itm:positive_feedback} of the explanation presented in Section~\ref{sec:industrialMRS}, and an absence of \emph{skip} (the music was entirely played) as a positive signal, such as in part \ref{itm:negative_feedback}, even though the music may have been played without someone actually listening to it.

Some design choices can also be made in order to make the system computationally efficient, notably limiting the amount of data: for instance, in item \ref{itm:history} of Table \ref{table:transparent_explanation}, the system needs to explain that old interactions were not taken into account for providing the recommendation, otherwise a user may not understand why some recurrent \emph{skip} of an artist they dislike was not taken into account. It is worth noting that such design choices are usually optimized in the industrial context (\eg through A/B testing), but are rarely considered in academic research.

We could also mention that it is common, in large catalogs, to encounter metadata ambiguities such as homonym artist profiles or polysemous musical genres. The impacts of such ambiguities on the system can be large and put explanation at risk of being deceptive, for instance, if an example-based explanation "Because you listened to artist X" is displayed to a user that listened to a different artist named X.

In the artificial explanation presented in Table \ref{table:transparent_explanation}, items \ref{itm:positive_feedback},  \ref{itm:negative_feedback}, and \ref{itm:history} rely on pragmatic assumptions.
This makes the explanation quite complex and may decrease user satisfaction, especially if some assumption is invalid: \eg when a song was skipped because played in an inappropriate context and not because it was disliked.
Furthermore, providing such a detailed explanation may change the user behavior \wrt these interactions: \eg they may avoid skipping songs they like in order to prevent them from being discarded in future recommendations, which may cause dissatisfaction.

%These assumptions may prevent providing clear explanations for a recommendation as the explanation would require giving insights about the engineering assumptions which could go against the needed simplicity of explanation.
%For instance, a recommendation may rely on a set of songs that were not skipped in a previous listening session of a user to provide similar songs in a new recommendation session. Then, to be sufficiently clear, an explanation of such recommendation should provide the following information: 
%\begin{itemize}
%    \item the assumption that non skipped songs were considered as somewhat liked by the user
%    \item songs similar to non skipped songs of a previous session were proposed
%\end{itemize}

\subsubsection*{Trade off between simplicity of the explanation and complexity of the \ac{RS}}
Among industrial actors, the \emph{less is more} design pattern is widely adopted as a general good practice supported by the theory of cognitive load applied to user interface design \cite{zhou2013affective}. The latter suggests that unnecessary display of information goes against ergonomics principles \cite{scapin1997ergonomic}, and can thus be detrimental to users' satisfaction. Following these guidelines, end-user explanations should be carefully crafted to remain simple and concise, hence making
cognitive overload less likely. 

Additionally, industrial incentives are primarily driven toward highly accurate systems.
%licity of explanations can clash with the complexity of the \ac{RS} generally considered by industrial actors as necessary to ensure high recommendation accuracy.
This often requires complex \ac{MRS} components,
%Several aspects of the \ac{MRS} design can bring complexity,
making explanations not simple enough to be provided to the user.
For instance, some constituting blocks can be based on black-box methods, % (see Section~\ref{sec:example_based_prototype})
such as latent factor-based models or deep embeddings that are widely used in \acp{MRS}. These models embed users and items as multidimensional vectors in a latent space, and represent affinity as their relative distance %(or any other notions of vector similarity) 
in this space. While they usually provide good results in a large variety of recommendation tasks, the latent factors are very difficult to understand: for instance, in item \ref{itm:similar} of Table \ref{table:transparent_explanation}, artist similarity is computed by a black-box system which is barely explainable to the user.

Besides, % as presented in Section~\ref{sec:industrialMRS}, 
several \ac{MRS} processing blocks rely on parameter choices. For instance, one may not consider all past user-song interactions to train the user-item affinity model, but only those that are significant (\eg only consider interactions when the user listened to at least half of a song). But this threshold is arbitrary and may exclude interactions that are important to a user: \eg users only listening to the intro of a song many times because they like it a lot. Arguably, these parameters should be optimized, but in practice they are so numerous that optimization becomes intractable.

Finally, %as stated in Section~\ref{sec:industrialMRS}, 
%the recommendations made by an
industrial \acp{MRS} are built upon several sub-blocks that are glued together and that rely on various sources of data: user modeling, content modeling, user-item affinity modeling, etc. The recommendation is made on top of all those blocks that may each influence the final recommendation. The impact of each block on this final recommendation is quite difficult to assess and, consequently, it is hard to generate a simple explanation on top of these unclear impacts. Feature selection may appear as a solution, but as long as several features are significantly impacting the prediction, the explanation would need to be either complex or incomplete. The overall complexity of explanations in Table~\ref{table:transparent_explanation} illustrates this issue.

\subsubsection*{Issues of transparency with respect to company competition}

One of the main goals of explanations for \acp{RS} is to increase transparency. While transparency can boost user satisfaction, it can possibly disclose some critical aspects of the system. Therefore, making sure that explanations do not reveal insights about the system internals can be necessary.
For instance, releasing the information that the \ac{MRS} uses artist embeddings (item \ref{itm:embeddings} of Table \ref{table:transparent_explanation}) or a specific hyperparameter of the system such as the considered time-frame in the history (item \ref{itm:history}) can be sensitive information that a private company may be reluctant to make public to competitors.

\subsection{Perspectives for explainable MRSs}

Improving the level of explanation of \acp{MRS} while keeping strong simplicity constraints for the user remains a challenge. Though, the end-user is not the only stakeholder to be impacted by \acp{MRS}. For instance, the revenue of music producers is impacted, too. Global explanations % mentioned in Section~\ref{sec:local_global}
may thus be relevant for such an audience, in terms of fairness and transparency (explanations would not be about single recommendations but probably about explaining why an artist was recommended to a particular group of people). As there are no simplicity constraints for this kind of stakeholder, explanations could possibly be much more elaborate.

Another aspect is that keeping a system explainable is important for constantly improving its performance. For instance, receiving user complaints or feedback about bad recommendations can only be leveraged for improving the system if the \ac{RS} engineers can understand the reason for these mis-recommendations. A \ac{RS} that relies on black-box blocks prevents understanding bad recommendations and, therefore, hinders improving the system.
%Marking that A \ac{RS} relies on black-box blocks may prevent understanding bad recommendations and thus hinder improving the system.

Finally, %it is worth noting that 
advanced users may want more control, and %that the 
simplicity constraints may be less important to them: For instance, \cite{jin2017different} %seems to 
argues that, as opposed to the \emph{less is more} design pattern, giving users additional control over the \ac{RS} does increase cognitive load, but also increases user satisfaction for users who have a deep understanding of how the \ac{RS} works. Controls that would enable interacting with the \ac{MRS}, make possible a positive feedback loop: explanations can be explicitly leveraged by the user to act on the \ac{RS} and mitigate future spurious recommendations.

Interestingly, the increasing usage of voice-controlled devices to pilot music streaming services creates a promising new playground for deploying explainable \acp{MRS} and beyond, to create fully interactive experiences where recommendations can be challenged, and eventually improved.

\section*{Acknowledgments}
This work received support from the Austrian Science Fund (FWF): P33526 and DFH-23.

\bibliography{biblio.bib}
%\bibliography{short_bib.bib}
\bibliographystyle{apalike}

\end{document}